2006 CCRTS
The State of the Art and The State of the Practice

Reading the Mind of the Enemy:
Predictive Analysis and Command Effectiveness

Topics:
C2 Analysis
C2 Experimentation
C2 Modeling and Simulation
Cognitive Domain Issues

Michael Ownby, Dr. Alexander Kott

Point of Contact
Michael Ownby
Solers, Inc.
1611 N. Kent St
Arlington VA 22209
571-218-4272, F703-741-7847
michael.ownby.ctr@darpa.mil

# Abstract

The Defense Advanced Research Projects Agency (DARPA) Real-time Adversarial Intelligence and Decision-making (RAID) program is investigating the feasibility of "reading the mind of the enemy" – to estimate and anticipate, in real-time, the enemy's likely goals, deceptions, actions, movements and positions. This program focuses specifically on urban battles at echelons of battalion and below. The RAID program leverages approximate game-theoretic and deception-sensitive algorithms to provide real-time enemy estimates to a tactical commander. A key hypothesis of the program is that these predictions and recommendations will make the commander more effective, i.e. he should be able to achieve his operational goals safer, faster, and more efficiently. Realistic experimentation and evaluation drive the development process using human-in-the-loop wargames to compare humans and the RAID system. Two experiments were conducted in 2005 as part of Phase I to determine if the RAID software could make predictions and recommendations as effectively and accurately as a 4-person experienced staff. This report discusses the intriguing and encouraging results of these first two experiments conducted by the RAID program. It also provides details about the experiment environment and methodology that were used to demonstrate and prove the research goals.

## Introduction

The Information Exploitation Office (IXO) of the DARPA is conducting a research project titled, RAID [1,2], to investigate the feasibility of "reading the mind of the enemy" – to estimate and anticipate, in real-time, the enemy's likely goals, deceptions, actions, movements and positions. A particular focus of the program is tactical urban operations against irregular combatants – an especially challenging and operationally relevant domain. The predictions and estimates can be provided to a commander via simple electronic map overlays. Given these timely predictions and estimates, the commander should then be able to conduct and modify his movements and tactics (course of action) during execution to achieve fewer casualties, faster completion of tasks, and more efficient use of resources. The development schedule for the RAID program is three 1-year phases. Two experiments conducted in 2005 measured the Phase I progress toward improving command effectiveness and yielded some intriguing and encouraging results regarding our ability to anticipate and predict a live enemy force in an urban environment. There is statistically significant evidence that the RAID predictions are as accurate and as effective as (and even on average more effective than) those of a 4-person experienced staff. When given RAID predictions and recommendations, the commander was more effective in accomplishing his given mission as measured by operational metrics, such as time to complete, blue and red casualties, and other mission-specific parameters.

The RAID program leverages approximate game-theoretic and deception-sensitive algorithms to build predictive analysis tools that provide real-time enemy estimates to a tactical commander. In doing so, the RAID program is addressing two critical technical challenges: (a) adversarial reasoning: the ability to continuously identify and update predictions of likely enemy actions [3]; (b) deception reasoning: the ability to continuously detect likely deceptions in the available battlefield information. Although many types of military operations can greatly benefit from these capabilities, the RAID program is focusing on an intentionally narrow but still very challenging domain: the in-execution, tactical combat of largely dismounted infantry (supported by armor and air platforms) against a guerilla-like enemy force in an urban terrain. Realistic experimentation and evaluation are driving the development process using human-in-the-loop wargames to compare humans and the RAID system. The Army OneSAF Testbed (OTB) wargame is the combat simulation system and is operated with live players controlling both the enemy (red) and the friendly (blue) forces. The products being developed by the RAID program are being evaluated for potential transition to battle command systems and military intelligence systems.

This paper discusses the motivation for the experiments and then provides a system overview followed by an explanation of the experiment design and operational scenarios used to measure the capabilities of the RAID system. Also provided is a discussion of the specific output products of the RAID system, such as predictions, recommendations, estimates of enemy goals and attitudes, identification of deceptions, anticipation of enemy movements, and suggestions for friendly countermoves. Finally, this paper reviews the outcome of the experiments and provides insights into the development and transition plans of the program.

## Motivation and Objectives of the Experiments

The objective of the RAID experiments is to explore the ability of RAID to make effective estimates of enemy actions and assumptions about friendly counteractions (move-countermove reasoning), as compared to a human staff. The experiments are not meant to replicate any existing staffing structure or command organization, but to measure the predictive capability of the software algorithms and compare the products to the predictive capability of a very experienced group of soldiers. In the experiments, RAID performs both intelligence and operational functions. First, it reads the enemy/friendly situation from the Combat Simulation System. For these Phase I experiments, situational information, such as location, strength, orientation and movement, was available in real-time and was 100% accurate and complete. Second, it accepts guidance from the blue commander (priorities, key objectives, etc.) and uses that information in its calculations. Third, it estimates, on demand, the most effective actions of enemy and assumed actions of blue for the next X minutes of wargame time. For these Phase I experiments, the estimates were for the next 30 minutes. Fourth, it completes every new estimate rapidly. In fact, portions of the algorithms run continuously in the background and give alerts as needed. Finally, the RAID system presents the estimate to the blue commander as overlay graphics or two or three dimensional (2D/3D) animations. The human staff performs these same functions and shares their data through an electronic whiteboard.

In addition to the comparative analysis, another objective of the experiments is to measure the performance parameters of the system. The specific parameters and goals of interest for each phase of the program are shown in Table 1. These first two experiments used Phase I goals. The look ahead into the future was a continuum of position, movement and firing predictions from 0-30 minutes. The problem complexity was calculated to be $10^{**}10,700$ for the types of scenarios and missions used in these experiments. For solution speed, the maximum time needed to calculate a new prediction was 300 seconds and the average was a 120 seconds. For the key goal, RAID had a higher average operational score than the human staff and beat the human staff in 11 of 18 paired runs. The RAID system was also more accurate in predicting the future locations of enemy forces and faster in identifying operationally pertinent phenomena than the human staff.

|  | Phase 1 | Phase 2 | Phase 3 |
|---|---|---|---|
| Look Ahead Into Future | At least 30 min | At least 60 min | At least 5 hours |
| Problem Complexity | over 10**8,000 | over 10**20,000 | over 10*50,000 |
| Solution speed | Within 300 sec | Within 120 sec | Within 30 sec |
| Key Goal | RAID-assisted small staff scores as high as large unassisted | RAID-assisted small staff scores as high as large unassisted | RAID-assisted small staff scores as high as large unassisted |

Table 1. RAID Program Experiment Goals by Phase.
Phase I was completed in 2005 and Phase II development is in progress. In addition to the increasing difficulty of the metrics in each phase, the scenarios and operational parameters also become more realistic.

Overview of RAID System

A simplified block diagram of the RAID system is depicted in Figure 1 showing the two major components of the RAID system – the Adversarial Reasoning Module (ARM) and the Deception Reasoning Module (DRM). Also shown is the Combat Simulation System which is used to simulate the battlefield and the information flows. There are two major inputs to the RAID system, battlefield models and situational information. The battlefield models include internal models of the terrain (3-D terrain data to include doors, windows, floors, and other details of the landscape); internal models of enemy and friendly resources (weapons parameters to include range, lethality, probabilities of acquisition/kill/damage/injury and vehicle or platform parameters to include min/max/avg speeds, vulnerability to specific weapons); and internal models of tactics and courses of action (to include influence of goals and objectives (speed, safety, success) on the selection of tactics). Situational information includes enemy and friendly force strength, location, and movement and is comprised of intelligence data from both Intelligence Preparation of the Battlefield (IPB) and updates from deployed sensors and operational data from both mission planning and operational reports from the deployed forces. Because the RAID system is designed for a future battlefield environment, it expects the situational data in a machine-readable format and does not work with voice or text-based sitreps, raw sensor feeds, or other unique reporting formats. For experimentation purposes, there is a translator between the Combat Simulation System and the RAID system which filters and translates the simulation data to provide the desired data stream to the RAID components. The output of the RAID system consists of enemy predictions and friendly recommendations and are discussed in detail later.

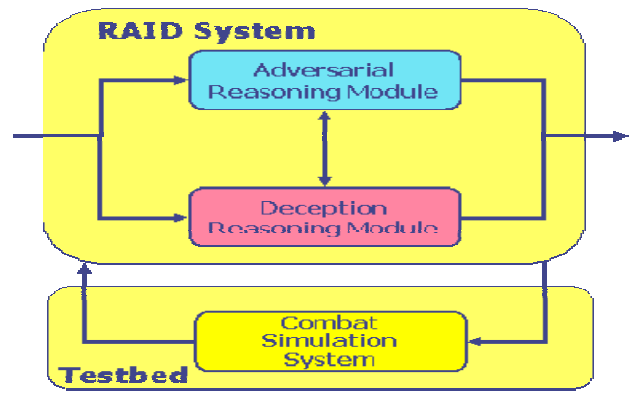

Figure 1. RAID System Diagram.
The two major components of the RAID system are the Adversary Reasoning Module and the Deception Reasoning Module. Although, the Combat Simulation System is not a part of the RAID system, it is integral to the development and testing.

The ARM is responsible for generating predictions of the enemy actions and assumptions about the friendly actions, either on-demand or in response to battle situation changes. As information regarding the battlefield situation (locations, strengths, postures, actions, etc.) of enemy and friendly troops becomes available or changes, either in the deliberate IPB and wargaming mode or during the execution of the operation, the ARM generates a new or modified set of predictions, including most dangerous and most likely prediction, each characterized by its likelihood. The ARM has two major subcomponents: one focuses on predicting the emotional state and intent of the enemy and the other focuses on the physical locations and movements of the enemy.

The DRM is responsible for identifying probable enemy deceptions, decoys, feints, and concealed enemy assets, movements and actions within the currently available information. The DRM provides the ARM with probable locations and concentrations of unseen enemy assets for use in the ARM calculations and provides alerts to the user or commander regarding decoys, feints or possible ambushes that represent near-term threats. While continually observing the evolution of the battlefield and the evolution of the predictions made by the ARM, the DRM continually updates itself and infers possible concealed enemy force elements or movements of elements, incorrectly identified enemy assets, decoys, actions designed to mislead friendly forces, etc.

The core of the experimentation testbed is the Combat Simulation System, which is based on the proven Army simulation and training system, OTB. Certain modifications to the existing system's interfaces and entity behaviors have been implemented to meet the needs of RAID experimentation. Both friendly and enemy teams of entities are controlled by human 'players' that can freely control movement, placement, and orientation. For the most part entity control is at the team level and not at the individual entity level. Fires are automated actions with the simulation system calculating the line-of-sight to determine if weapons are fired and using probability tables to determine degree and amount of effect or casualties. Although the simulation system is not a part of the RAID system under development, it is a critical component for exercising the operational functionality of the RAID components. The simulation system is used by the technology developers as a means of observing and learning both friendly and enemy capabilities and tactics. Finally, the simulation system is used to measure and demonstrate the performance characteristics of the RAID system.

**Experimental Methodology**

The basic premise of the RAID experiment methodology is to compare the performance of a commander using RAID (test games) against a commander without RAID, but supported by a small staff of human experts (benchmark games). A tactical commander at this level of command (below battalion) would not typically have a dedicated staff to help him during the real-time execution of an urban battle, but in order to measure the accuracy and value of the RAID system, the experiment commander is provided with a staff of two operational experts and two intelligence experts. During the benchmark games, this staff provides the same kind of predictive products that are produced by RAID. Thus comparative measurements can be made with regard to prediction accuracy, timeliness, and completeness as well as benefit to the commander's success (mission accomplishment). Each game lasted about 2 hours (preceded by an orientation and planning session) and Figure 2 illustrates the setup used for the experiments. Forces were tasked and controlled as teams and not as individual soldiers. The wargame players each controlled 5-6 teams and used tasks (which translate to a series of automated behaviors in the simulator) to move and command their forces. The friendly forces had five Strykers, which were managed by one player and controlled individually.

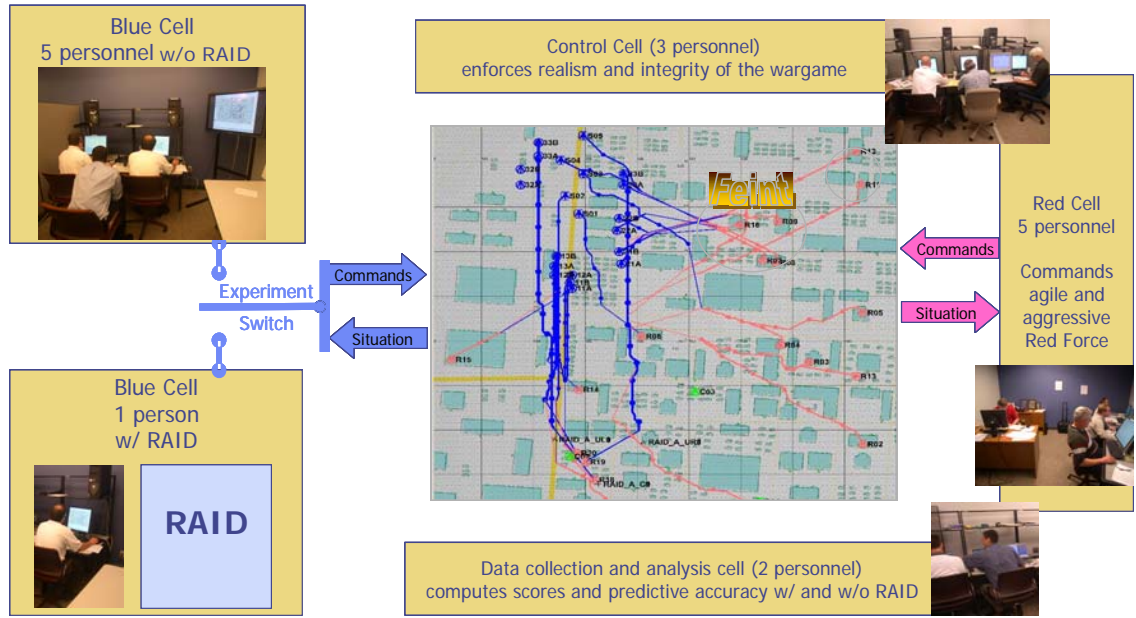

Figure 2. Experiment Setup.
This shows the physical setup for the experiments, with a Blue command cell, a Red cell playing the enemy forces, a White cell ensuring integrity of the wargames and paired runs, and a data collection cell gathering metrics and other information.

Control of the enemy team entities was performed by a red cell of five experienced human wargamers who are trained in the typical tactics of urban insurgents. The red cell consisted of a red commander and four operators who each controlled 5-6 red teams. Each team had three individual entities. The red cell could move their teams anywhere and employ any tactics to achieve their objectives or thwart the friendly objectives. Also, the red cell was not told if they were playing a benchmark game (against the human staff) or a test game (against the RAID system). Exit interviews with the red cell confirmed that they were not able to discern whether their opponent was human or software.

Planning and overall control of friendly forces was performed by a blue command cell, as depicted in Figure 2. However, control of the actual friendly team entities in the simulation was by a blue cell of 4 humans who received their plans, directions and commands from the blue commander. The blue commander used the predictions and recommendations from the RAID system or his "staff" to develop the course of action. For the Phase 1 experiments, the battle was paused every 15 minutes and the commander received an updated set of predictions and recommendations. Throughout each game, the commander was free to request new updates, as desired, and could alter his course of action at will. The blue commander was rotated for various scenarios, but the same commander was used for both the benchmark and the test game of each paired set of runs. To guard against learning, the paired games were separated by several days;

sometimes the benchmark version was played first and sometimes the test version was played first. In either case the commander was not told or reminded whether he was playing the first or second half of the pair of games. Great care was taken to ensure the integrity of the paired games and during both Phase I experiments a few pairs were thrown out due to contamination of the process.

As part of the white cell control process, both the red and blue commander provided the white cell with a general overview of their planned course of action for a given scenario after they had a chance to review the initial laydown of forces and were given their mission objectives. Certain red teams were assigned emotions to play and were required to log the actions they took to represent those emotions. This provided the data collectors with ground truth regarding the emotional states that might be identified during the battle. Finally, all communications were monitored to ensure the integrity of the wargames.

**Experimental Scenarios**

Scenarios for the experiments were focused on an urban terrain with a largely dismounted Blue Force operating against an insurgent-like irregular dismounted Red Force. The scenarios were inspired by recent military engagements in Mogadishu, Najaf, Fallujah, and other urban battles. The situations emulated included the defense of friendly government facilities, the rescue of downed aircrew, the capture of an insurgent leader, the rescue of hostages or the reaction to an attack on a friendly patrol. For operational simplicity, only a few basic mission types were used to build the scenarios for the experiments. The three basic mission types were Point/Area Attack (blue forces attack a specific target or area, Point/Area Defense (blue forces protect specific buildings and clear all enemy forces from a specific area) and Withdrawal (congregate all blue forces at a designated location). Figure 3 illustrates a typical starting laydown of forces for a

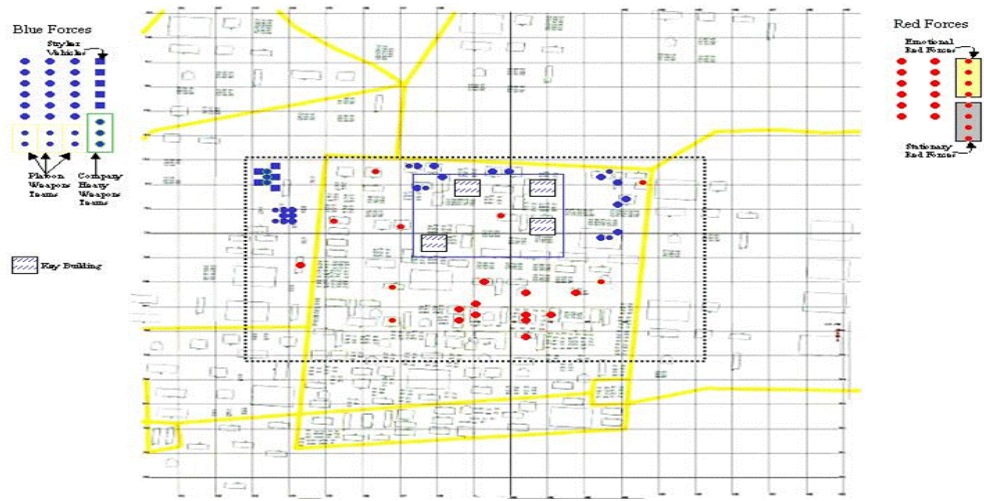

Figure 3. Example of Point Defense Mission.
In a point defense mission, Blue Forces were tasked to protect specified buildings from attack. The operational score was influenced by how quickly the blue forces reached the objective buildings and how well they protected them.

Point Defense scenario where friendly forces are tasked to protect the four buildings highlighted in the center. The enemy's objective was to attack any number of those four facilities to inflict casualties and damage. In this type of scenario, blue forces would need to move as quickly as possible into defensive positions around the designated buildings. As can be seen, some enemy forces are initially closer to the buildings of interest and could cause damage and casualties while blue forces were enroute. The operational scoring for the point defense mission was greatly influenced by how quickly blue forces were in place around the designated buildings and how much (how little) fire the buildings received from the red forces. As an example of the free play exercised by the red forces, sometimes they attacked all the target buildings, and sometimes they focused their efforts on only one or two. Likewise, the blue commander would typically try to protect all the buildings, but would occasionally sacrifice (leave unprotected) a single building to better protect the remaining buildings.

For the experiments discussed in this report, the scenarios also contained the following characteristics:
a. the friendly forces were comprised of 18 teams (each w/4 dismounts armed with M-16s) and 5 Strykers (armored attack vehicles). For battle control purposes, the 18 teams were organized as 3 platoons of 6 teams each with one simulation operator controlling each platoon. The Strykers were assigned to platoons, but were controlled by a Stryker operator, not the platoon operator.;
b. the enemy forces consisted of 20 teams (each w/3 dismounts armed with AK-47s and one rocket propelled grenade (RPG) launcher). The RPGs were effective against the Strykers, but the AK-47s were not. Also, red forces were given slightly faster movement speeds due to their "familiarity" with the terrain.;
c. the battle area was a 2 kilometer (Km) by 2 Km region of an urban city with multi-story buildings (floors, doors, and windows were played in the simulation). Each game was played within a 1 Km by 1 Km section of the larger battle area.;
d. the scenarios were designed to take about two hours to complete the mission. Some of the scenarios were designed with two sequential missions, such as rescue a downed aircrew and then withdraw to a specified rally point.; and
e. for Phase I experiments only, full Intelligence was known to both sides with regard to location, strength, and movement of all forces (much like a chess game). This artificiality was used to prove that full knowledge of the battlespace (e.g. overwhelming intelligence, surveillance and reconnaissance (ISR) resources) is not sufficient to guarantee success in battle and it made helped reduce some of the complexity for the immature, evolving technologies. In subsequent phases, both the initial Intelligence of the distribution of forces will be greatly diminished (to approximately 25%) and the ongoing updates during the execution of the simulation will be severely constrained to more accurately replicate real world conditions.

The experimental scenarios and methodology have been setup to provide sufficient military relevance to prove the value of the technology to a military audience and designed with sufficient complexity to prove that the technology can handle the extraordinary complexity of real world battles. To reiterate, the experiments were not been setup to replicate any current or future command and control (C2) or intelligence process, structure, or organization, but have been designed to measure the accuracy, timeliness, and value of the predictions and recommendations of the RAID system as compared to what a group of highly experienced humans could do with the same information. The next section of this report discusses the output products produced by the RAID system.

## Predictions and Recommendations

The main focus of the RAID system is to make predictions about enemy actions, to include goals, deceptions, actions, movements, and positions. In the classic military move—countermove—counter-countermove planning and analysis, the RAID system makes assumptions or recommendations about friendly plans and behaviors to accurately predict the enemy. In formulating the predictions, the ARM takes into account such factors as the high-level objectives, intents and preferences of the friendly and enemy commanders, physical capabilities and needs of the assets available to both sides, mutual influence of actions of blue and red forces, terrain, non-combatants, cultural and doctrinal aspects, psychological factors affecting troops and commanders, prior evolution of the operation, etc. With this input information, the ARM generates a detailed prediction looking forward anywhere from 0 to 300 minutes (as specified by the user) from the current moment, including a sequence of actions (situated in time and space) to be performed by the enemy force. During Phase I, the RAID system only made predictions up to 30 minutes into the future.

Because the actions of red and blue forces are closely connected and influence each other, the ARM must also generate its estimates of the friendly actions similar to the predictions of enemy actions. These can be seen as assumptions or recommendations regarding the friendly course of action. Although the primary function of RAID is to anticipate red actions, the capability to suggest friendly actions is a natural, valuable byproduct that can be effectively used by the commander via an integrated command and control/intelligence system. As an integral part of the RAID predictive process, the DRM infers possible concealed enemy force elements or movements of elements, incorrectly identified enemy assets, decoys, actions designed to mislead friendly forces, etc. by continually observing the evolution of the battlefield and the evolution of the predictions made by the ARM. Since there were no concealed forces during the Phase I experiments, the DRM only made predictions about potential feints or misdirections attempted by the enemy.

## Estimation of Enemy Goals and Attitudes

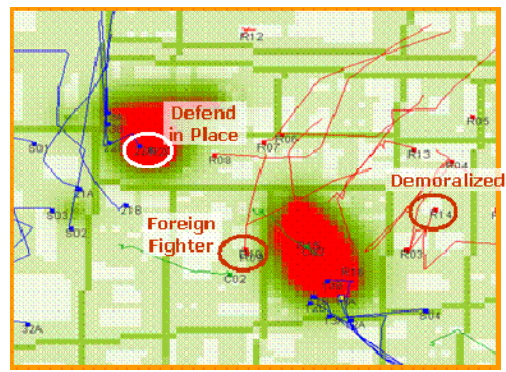

Figure 4. Cognitive Predictions. The ARM can predict the goals and attitude (emotional state) of the enemy.

One portion of the ARM has been designed to explicitly handle the "human" aspects of battlefield behaviors with a cognitive model (Bayesian belief net) that propagates relations between actions, emotions, goals, desires and dispositions. Specifically, this sub-component captures the implicit cultural and doctrinal preferences and then connects the observed behaviors with an estimated mental state. The projected mental state is then mapped into probable incipient goals. As the enemy takes actions and moves about the battlefield, a pheromone-analogy algorithm fits these current behaviors to past behaviors and prunes and clarifies its model of the mental state of the enemy. Finally, this sub-component projects future "broad-brush" physical behaviors and mental state evolution by exploring multiple potential roll-outs of actions and events using a concept of ghost agents or avatars [4,5,6,7,8].

Figure 4 illustrates a few of the predictions made by this sub-component of the ARM. Predictions of emotional states were displayed as 'alerts' as they were identified. Identification of the emotional state of an enemy unit (team) can be beneficial to the commander in determining if that enemy team is an actual threat that needs to be countered or merely a nuisance that needs to be monitored. The prediction of enemy goals helps the commander to understand the prediction of enemy movement. In this example, the enemy is predicted to 'defend in place' with an accompanying movement prediction that the enemy will not advance, but will be in a defensive posture. Finally, the prediction of 'foreign fighter' will give the commander a sense of the motivation and zeal of that enemy team as well as possible insight into the potential skill level or lethality of that team. For example, during one mission the RAID system identified a "cowardly" fireteam on the flank of the blue forces. Using that prediction, the commander chose not to reinforce his flank and to continue his forward motion to the objective. As predicted, the enemy fireteam never attacked the exposed flank of the blue force. During the second Phase I experiment, the human staff was asked to identify the demoralized or unenthusiastic enemy fireteams and report them as soon as they were identified. Figure 5 compares the detection/predictions from RAID versus the human staff. In general, the human staff and the RAID system identified the same "emotional" fireteams, but the RAID system consistently made the identification 10-15 minutes sooner than the human staff. The faster identification times could make a difference in the tactical strategies employed and ultimate outcome of the engagement.

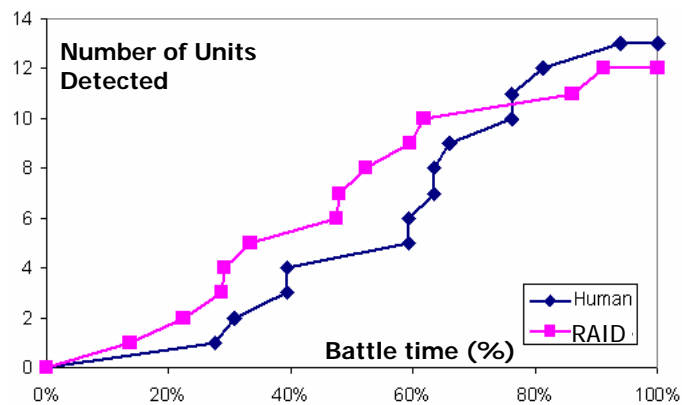

Figure 5. Prediction of Enemy Emotions. The RAID system and the human staff were able to detect and predict the same set of "emotional" enemy fireteams, but the RAID system made the identification 10-15 minutes faster than the staff.

**Identification of Enemy Deceptions**

The DRM is responsible for identifying probable enemy deceptions, decoys, feints, and concealed enemy assets, movements and actions within the currently available information. It employs multiple technologies, such as: a deception robustness estimator that applies stochastic game theory to state estimation to discern underlying deception strategies [9,10,11,12]; a deception cost/value evaluation which combines consideration of observations, cost for Red to deceive, and value to Red if deception works; a risk-sensitive theory for recognition and analysis of deception potentials and likelihoods; and non-symmetric evaluation functions that produce value functions, initially through SME heuristics, then through automated learning.

Figure 6 illustrates a 'Feint' alert displayed by the DRM component during the Phase I experiments. Due to the full information state, there were no hidden forces or concealed movements for the DRM to predict, so feints were the only information predicted by the DRM. During the Phase I experiments, the Red cell knowingly attempted nine feints. The RAID system identified four and the human staff identified three (see Figure 7). Although this is not a statistically significant difference, it is encouraging that software was able to discern such a subtle behavior anomaly. In the next phase of development, the DRM will be exercised fully with limited information about an enemy that can easily conceal its assets, movements, and actions. Also, the prediction of enemy locations by the DRM will become a necessary and important input to the ARM component.

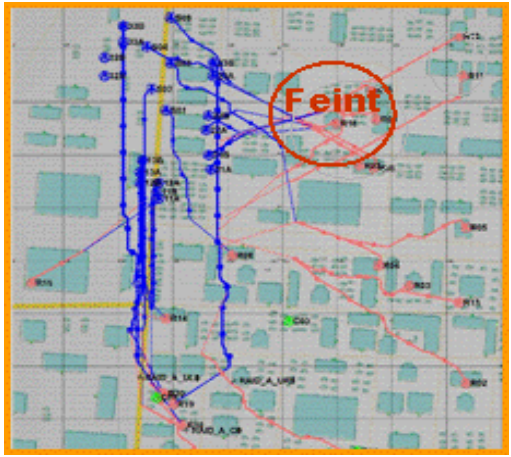

Figure 6. Deception Prediction
The DRM predicts a feint at this
location during the Phase I experiments.

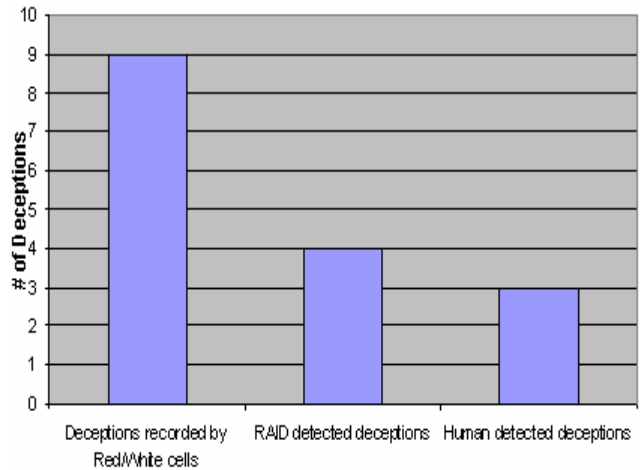

Figure 7. Feint Prediction.
RAID was able to detect and identify a higher
percentage of enemy feints than the human staff.

## Anticipation of Enemy Movements

The other sub-component of the ARM predicts enemy movements and recommends friendly courses of action to best counter the enemy. This prediction technique employs an efficient abstraction (known as Linguistic Geometry)[13,14] of the action space for a non-zero-sum game solution. This technology uses a small number of general-purpose heuristics to guide a fast, low-branching search routine, then generates multiple worldviews to reflect the partial observability of Red and Blue, and finally provides a means to view the predictions and recommendations in both 2D and 3D as an animation over the prediction time window. Figure 8 illustrates a snapshot from both views. The red lines represent predicted moves by the enemy forces and the blue lines are the recommended moves for the friendly forces. The illustration also shows where engagements (fire fights) are likely to occur and the 3D version can even show forces using upper floors and even rooftops for better coverage and fields of fire. The animation adds timing information to the predictions to allow a commander to see and understand the predicted sequence of movements and to better understand the necessary synchronization of his own troop movement. The 3D view also helped commanders appreciate the line-of-sight parameters of the battlefield, especially since the simulation terrain was not flat and allowed interesting line-of-sight opportunities.

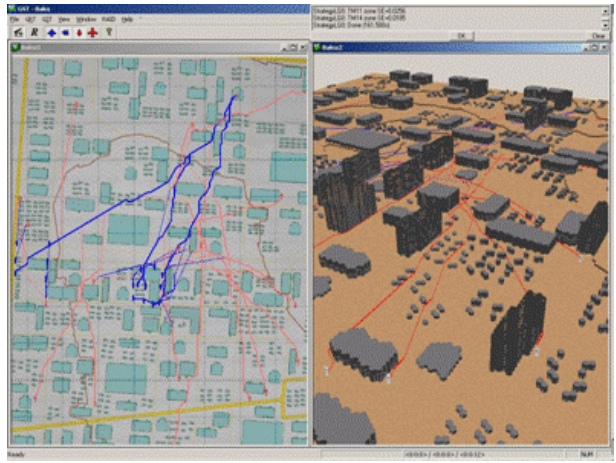

Figure 8. Prediction of Enemy Movements.
The left side is a 2D presentation of the predicted
moves and the right side is a 3D version. Also
shown are the recommended blue movements
and the predicted engagement zones.

In this area of movement predictions, the RAID system performed significantly better than the human staff (see Figure 9). On average, the distance error for the human predictions was twice the distance error for the RAID predictions. Both humans and the RAID system made better predictions as the battle progressed and end objectives (destinations) and initial movement directions became visible. For some games, the RAID system was able to correctly predict the movement of 19 of the 20 enemy fireteams within one building – meaning a fireteam went around a building on one side and RAID predicted the other side.

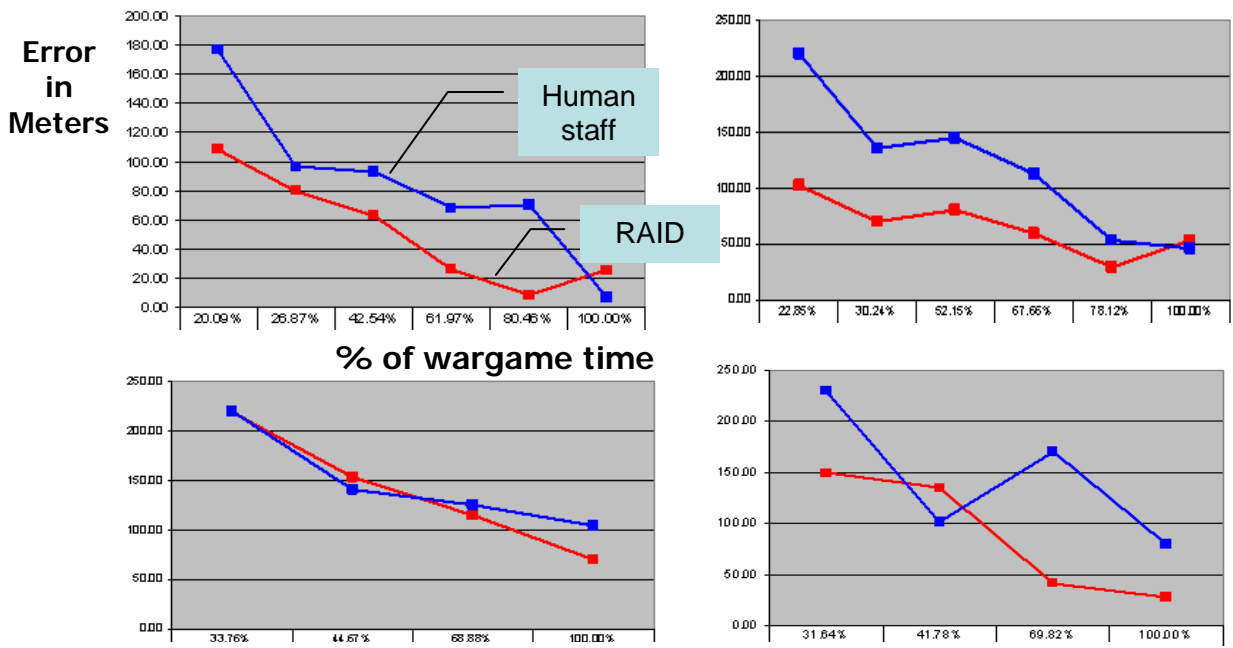

Figure 9. Accuracy of Predicted Enemy Locations
This is a sampling of the error (in meters) of the predicted enemy locations.
Generally, RAID performed significantly better than the human staff.

## Overall Battle Outcome

In addition to comparing the component level products of the RAID system with human experts, each game was scored using operational metrics to measure the success of the blue force, i.e. command effectiveness. The blue force command effectiveness was measured by: the rate of progress toward the mission accomplishment (e.g., advancing to or clearing the specified objective); the number of red personnel killed; and the avoidance of friendly losses and collateral casualties. Success of the red force was measured by delay of the blue force and causing blue casualties. To ensure that RAID supported runs truly represented the RAID predictions and recommendations, the blue commanders had to follow the general recommendations provided by RAID and deviate only as needed to implement the desired actions. Elements of the weighted scoring algorithm are shown in Figure 10. The weightings were tuned for the various mission types and reviewed by operational experts to verify the appropriateness of the values. Values were measured in real-time during the experiment runs and the Control Cell and Data Collection Cell were able to correlate value changes with scenario events for post-run analysis.

The chart in Figure 11 shows the outcome of the RAID supported runs versus the human supported runs in the second experiment. Similar values were obtained for the first experiment as well. Each symbol represents a paired run (where the mission and initial laydown of enemy and friendly forces were identical) with the RAID score on the vertical axis and the human score on the horizontal axis. Symbols that lie above the diagonal line are pairs where RAID had a higher score. In both Phase I experiments, the RAID supported runs had a higher average score than the human supported runs. For the nine paired runs, the RAID system had an average score of 75.5 and the human staff had an average of 72.3 with a standard deviation of 10 points. The statistical significance of the values was 81%. In exit interviews with the Blue commanders, their command effectiveness was better with RAID due to both more accurate predictions of enemy movements and details about the blue course of action that helped to convey the importance of timing and synchronization.

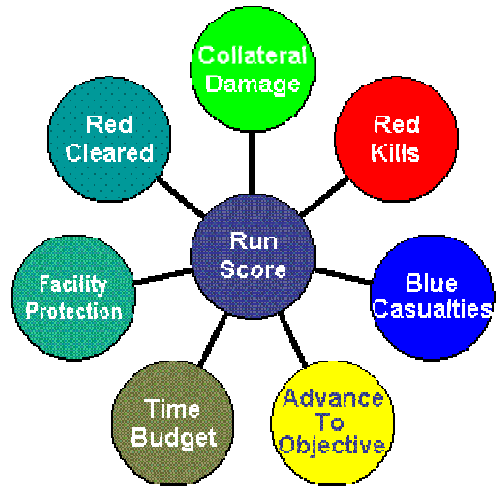

Figure 10. Measurement of Command Effectiveness.
This is a display of the weighted measures that are used to calculate the operational score for each run.

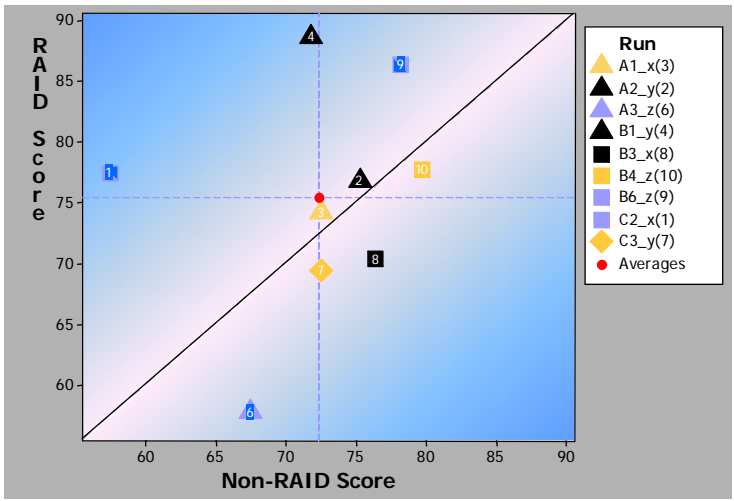

Figure 11. Summary of Scores.

This figure shows the paired values of nine experiment run pairs, clearly indicating that the RAID supported runs had a higher average

## Conclusions and Future Work

The experimental results are intriguing and encouraging. To think that a computer system can "read the mind of the enemy" and make meaningful and correct predictions about the enemy's likely goals, deceptions, actions, and movements is quite an accomplishment. To be able to do this in real-time against an intelligent, live enemy is almost unheard of. Exit interviews with the human staff and human players indicate that the RAID system does a better job of evaluating the entire battlefield and is not distracted by singular events. The RAID system did not forget, did not overlook, and kept track of all fireteams, which allowed it to make better predictions and recommendations. Also, the products produced by the RAID system provided a clearer understanding of the enemy, the enemy's most likely actions, and the most appropriate friendly counter-moves.

The RAID program is being conducted in three 12-month phases.
   a. Phase I – Adversarial Anticipation and Counteraction focused on mechanisms to compute and anticipate adversarial, move-countermove actions. Phase I is complete.
   b. Phase II – Adversarial Reasoning about Concealment and Deception will focus on the ability to see through the fog of war and recognize deceptions.
   c. Phase III - Integration and Transition to the Army's Distributed Common Ground System (DCGS-A) will develop fieldable products which can integrate with existing C2 and ISR systems.

The major change for Phase II experiments will be the elimination of full knowledge of the battlespace. Both enemy and blue forces will have limited initial knowledge of each other and will have limited "sensor" capabilities to learn about the opposing forces. Complexity will also be increased by adding more forces, more weapon types, a larger play area, communication delays, and other factors which add to the realism of the simulation environment. Additionally, measurement techniques are being added to the Phase II experiments to try and ascertain "why" the RAID system is able to do better. Phase II experiments will be conducted at the Army's Battle Command BattleLabs (Fort Huachuca and Fort Leavenworth) and will use active duty personnel as the blue staff.